%% file: main.tex
\documentclass[runningheads]{llncs}

\usepackage{eccv}

\input{math_commands.tex}

\DeclareMathOperator{\proud}{ProUD}
\DeclareMathOperator{\dist}{dist}

\usepackage{enumitem}
\usepackage{eccvabbrv}

\usepackage{graphicx}
\usepackage{booktabs}
\usepackage{pdfpages}

\usepackage[accsupp]{axessibility}  

\usepackage[pagebackref,breaklinks,colorlinks,citecolor=eccvblue]{hyperref}

\usepackage{orcidlink}

\usepackage{hyperref}
\usepackage{url}
\usepackage{comment}
\usepackage{enumitem}
\usepackage{booktabs}
\usepackage{multirow}
\usepackage{graphicx}
\usepackage{adjustbox}
\usepackage{float}
\usepackage{array}
\usepackage{wrapfig}
\usepackage{algorithm}
\usepackage{algpseudocode}

\usepackage{dsfont}
\usepackage{amsmath,amssymb}

\usepackage{setspace}
\usepackage{caption}

\definecolor{mygray}{gray}{0.5}

\begin{document}

\title{Overcoming Data Inequality across Domains\\with Semi-Supervised Domain Generalization} 

\titlerunning{Overcoming Data Inequality across Domains with SSDG}

\author{Jinha Park$^{*}$\inst{1} \and Wonguk Cho$^{*}$\inst{2} \and Taesup Kim\inst{2}}

\authorrunning{J.Park et al.}

\institute{Department of Electrical and Computer Engineering, Seoul National University \\ \email{jhpark410@snu.ac.kr} \and
Graduate School of Data Science, Seoul National University \\ 
\email{\{wongukcho, taesup.kim\}@snu.ac.kr}}

\maketitle
\let\thefootnote\relax\footnotetext{*Equal contribution}

\begin{abstract}
  While there have been considerable advancements in machine learning driven by extensive datasets, a significant disparity still persists in the availability of data across various sources and populations. This inequality across domains poses challenges in modeling for those with limited data, which can lead to profound practical and ethical concerns. In this paper, we address a representative case of data inequality problem across domains termed Semi-Supervised Domain Generalization (SSDG), in which only one domain is labeled while the rest are unlabeled. We propose a novel algorithm, ProUD, which can effectively learn domain-invariant features via domain-aware prototypes along with progressive generalization via uncertainty-adaptive mixing of labeled and unlabeled domains. Our experiments on three different benchmark datasets demonstrate the effectiveness of ProUD, outperforming all baseline models including single domain generalization and semi-supervised learning. Source code will be released upon acceptance of the paper.

\end{abstract}

\section{Introduction}
In the realm of machine learning, the availability of extensive datasets has played a pivotal role in driving its advancements~\cite{sun2017revisiting, kaplan2020scaling}.
However, acquiring sufficient training data remains a challenge due to disparities in data accessibility across different sources and populations, an issue commonly termed as \textit{data inequality}.
The World Development Report by the World Bank~\cite{world2021world} underlines this problem, noting that developing economies frequently grapple with data scarcity stemming from insufficient infrastructure for data connectivity, storage, and processing. The deficiency also extends to the limited availability of human expertise and skilled labor in these areas.

Such data inequality not only presents practical challenges but also raises ethical concerns in the design and deployment of machine learning models. 
This issue is prevalent across various fields, particularly in the biomedical sector. A clear example is the data inequality across ethnic groups, which can lead to uneven model performance and global healthcare inequalities~\cite{gao2020deep, gao2023addressing}. 
For instance, recent statistics~\cite{guerrero2018analysis} reveal a severe imbalance; data from 416 cancer-related genome-wide association studies were collected from Caucasians (91.1\%), followed distantly by Asians (5.6\%), African Americans (1.7\%), Hispanics (0.5\%), and other populations (0.5\%). This represents significant data inequality across ethnic groups, particularly given that non-Caucasians constitute approximately 84\% of the world's population. This inequality can cause machine learning models to exhibit low accuracy and a lack of robustness for these underrepresented groups, potentially harming their healthcare outcomes~\cite{martin2019clinical, rajkomar2018ensuring}.

\begin{table}[t]

\caption{Examples of machine learning applications in various fields susceptible to data inequality problem across domains, where Semi-Supervised Domain Generalization (SSDG) can be applied.}

\label{table:0}
\centering
\begin{adjustbox}{width=0.9\textwidth}
\setlength\extrarowheight{2pt}

\begin{tabular}{c||c|c}

\hline
\textbf{Task / Application} & \textbf{Labeled Domain} & \textbf{Unlabeled Domain} \\ \hline \hline
\multirow{2}{*}{Biomedical Imaging} & Caucasians & Other ethnicities \\ \cline{2-3} 
 & Central hospitals & Peripheral hospitals \\ \hline
\multirow{2}{*}{Natural Language Processing} & English & Minority languages \\ \cline{2-3} 
 & Standard language & Regional dialects \\ \hline
\multirow{2}{*}{Autonomous Driving} & Urban area & Rural area \\ \cline{2-3} 
 & Typical weather conditions & Rare weather conditions \\ \hline
Agricultural Crop Monitoring & Developed countries & Developing countries \\ \cline{2-3} 
 (Using Satellite Images) & Commercial satellites & Non-commercial satellites \\ \hline
\end{tabular}
\end{adjustbox}
\end{table}

\begin{figure}[t]
\begin{center}
\includegraphics[width=\linewidth]{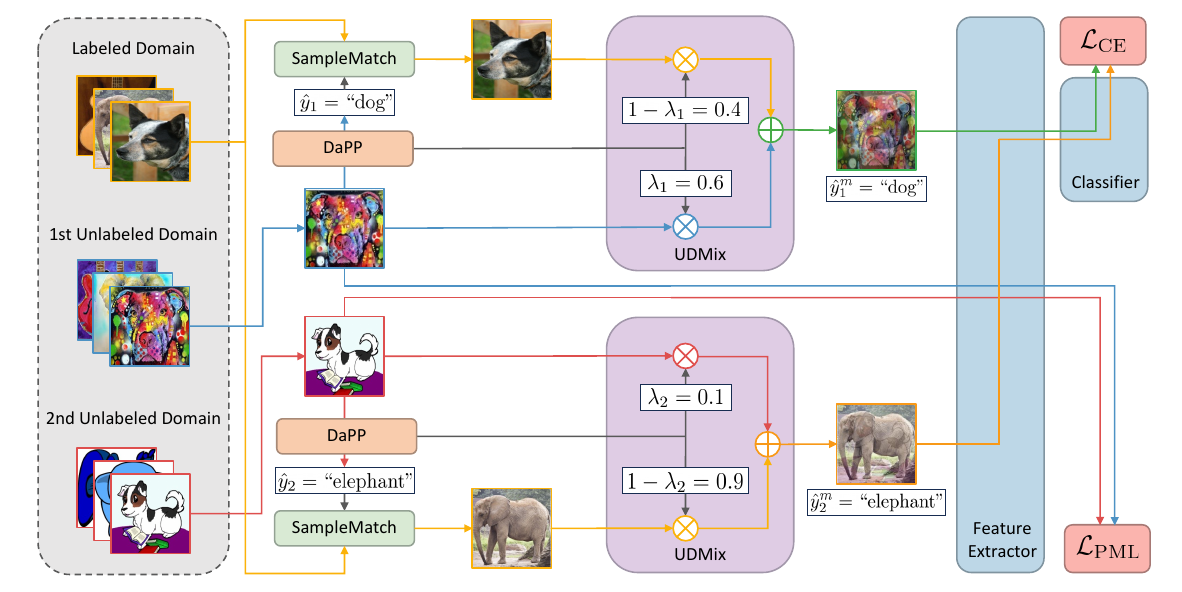}
\end{center}
\caption{ProUD training examples with three different domains: one labeled domain (highlighted in yellow) and two unlabeled domains (indicated in blue and red). For the blue unlabeled domain, DaPP accurately infers a correct pseudo-label and a substantial portion of the unlabeled domain image ($\lambda_1=0.6$) is mixed with the labeled domain image through UDMix. This results in a domain-mixed image depicted in green. In contrast, for the red unlabeled domain, an incorrect pseudo-label is generated, yet only a minimal fraction of this image ($\lambda_2=0.1$) is mixed with the labeled domain image through UDMix, leading to a predominantly yellow domain-mixed image.}
\label{fig:main}
\end{figure}

In light of these concerns, we address a representative case of the data inequality problem across domains, termed Semi-Supervised Domain Generalization (SSDG).
More specifically, the core objective of SSDG is to learn domain-invariant features from multiple source domains, wherein only one domain is labeled while the rest domains remain unlabeled. 
Such a setting mirrors real-world situations, especially when obtaining labeled data from certain domains is considerably more difficult than from others. 
Tab.~\ref{table:0} provides some examples of this scenario across various fields.

To address the SSDG problem, there exist two existing methods that can be applied: semi-supervised learning and single domain generalization. 
Semi-supervised learning is a method designed to train with a small amount of labeled data and a larger volume of unlabeled data. Nonetheless, this method assumes all training data originate from a single domain, thus it is not able to use domain information of training data in SSDG. This restricts its ability to learn domain-invariant features and generalize across domains. On the other hand, single domain generalization, which uses a single labeled source domain to achieve generalization, can be applied to SSDG in a way that exclusively uses the labeled domain among the source domains. However, its inability to make use of unlabeled domains may result in suboptimal performance, especially in domains that are not closely related to the labeled source domain.

To overcome these limitations of existing approaches, we introduce a novel algorithm called Prototype-based Uncertainty-adaptive Domain Generalization (ProUD), which is illustrated in Fig.~\ref{fig:main}. ProUD is specifically designed to leverage both domain information and unlabeled data. To effectively utilize domain information for learning domain-invariant features, we introduce domain-aware prototypes for each class. For the effective utilization of unlabeled domains, we employ uncertainty-adaptive mixing of labeled and unlabeled domains. This strategy modulates the involvement of each sample from various domains based on its uncertainty, thereby achieving robust and progressive generalization. The core idea and details of ProUD are further elaborated in Sec.~\ref{3}.

In Sec.~\ref{4}, we compare our method with extensive baselines including semi-supervised learning and single domain generalization methods on three benchmark datasets: PACS~\cite{li2017deeper}, Digits-DG~\cite{lecun1998gradient, ganin2016domain, netzer2011reading, roy2018effects}, and Office-Home~\cite{venkateswara2017deep}. Comprehensive experiments demonstrate that ProUD outperforms all baselines in terms of average accuracy and standard deviation across 12 distinct domain combinations in all datasets.

Overall, our contributions can be summarized as follows:
\begin{itemize}
\item We introduce a novel algorithm, ProUD, to address SSDG, a representative case of data inequality that mirrors real-world scenarios where obtaining labeled data from certain domains is considerably more challenging than from others. 

\item We develop two key methods to effectively utilize domain information and unlabeled domains: (1) learning domain-invariant features via domain-aware prototypes and (2) uncertainty-adaptive mixing of labeled and unlabeled domains.

\item We demonstrate the effectiveness of our approach by consistently achieving state-of-the-art performance on all benchmark datasets. Our method further achieves the lowest standard deviation over different domain combinations, highlighting its robustness.
\end{itemize}

\section{Related Work}
\subsubsection{Semi-Supervised Learning}
In semi-supervised learning (SSL), a small amount of labeled data is accessible alongside a larger volume of unlabeled data to train a model. Various methods have been developed to use this mix of labeled and unlabeled data. An established approach in SSL is consistency regularization, which forces a model's predictions to remain consistent when alterations are made to the model~\cite{laine2016temporal, tarvainen2017mean, miyato2018virtual}. Another popular approach is to employ pseudo-labeling~\cite{lee2013pseudo}, which generates pseudo-labels for unlabeled data using a pretrained model~\cite{xie2020self}. Furthermore, MixMatch~\cite{berthelot2019mixmatch}, FeatMatch~\cite{kuo2020featmatch}, and FixMatch~\cite{sohn2020fixmatch} incorporate a combination of pseudo-labeling and consistency regularization. 
However, typical SSL methods heavily rely on the assumption that labeled and unlabeled data share an identical distribution, which can be quite challenging to fulfill in real-world scenarios. SSDG can be viewed as a particular variation of SSL, specifically relevant in practical situations where the labeled and unlabeled data hold different distributions. The difference in distribution between the labeled and unlabeled data can result in significant bias in the pseudo-labels, leading to degradation in performance. To address this issue, EID~\cite{lin2024semi} filters out noisy labels using a specified cleaning rate to generate a set of pseudo-labeled data with enhanced quality, employing a dual network architecture. However, EID necessitates as many distinct DA models and training processes as there are unlabeled source domains, which severely restricts its practical applicability in SSDG scenarios involving numerous source domains. In contrast, ProUD, our progressive generalization algorithm, overcomes this limitation by simultaneously handling multiple unlabeled source domains with a single model, demonstrating its practical potential in addressing data inequality.  

\subsubsection{Domain Generalization}
Domain Generalization (DG) aims to train a model using data from one or multiple source domains, enabling it to achieve effective generalization across unseen target domains. Early research on DG primarily focused on acquiring domain-invariant representations by aligning features across distinct sources~\cite{gan2016learning,  ghifary2016scatter, ghifary2015domain, li2018domain}. 
Another approach to DG incorporates data augmentation, which seeks to increase style diversity at the image-level~\cite{zhou2020deep, xu2021fourier, gong2019dlow, zhou2020learning}, or the feature-level~\cite{zhou2021domaina, li2022uncertainty, zhong2022adversarial, li2021simple}.
However, these DG approaches are limited by their heavy reliance on accessing labeled data from multiple source domains, necessitating a discussion on their application in practical scenarios~\cite{cho2023complementary,simon2022generalizing,dubey2021adaptive}. To this end, single DG methods~\cite{romera2018train, volpi2018generalizing, zhao2020maximum} have emerged, which train with labeled data from a single domain. Single DG methods are largely built upon adversarial domain augmentation~\cite{huang2020self, qiao2020learning, wang2021learning}. Our problem setting, SSDG, is similar to single DG, as only a single labeled domain is available for training. However, it differs in that it also has access to unlabeled data from multiple domains. Our work focuses on effectively leveraging unlabeled source data by progressively mixing it with labeled source data through Uncertainty-adaptive Domain Mix (UDMix). Our experimental results confirm the significant advantage of leveraging unlabeled domains compared to using single DG methods.

\subsubsection{Mixup}
Mixup~\cite{zhang2017MixUp} is a simple yet effective method to extend the training data distribution, founded on the intuition that performing linear interpolations among input images will result in corresponding linear interpolations of the labels. As an intuitive strategy for data augmentation, mixup has also been studied in the context of DG~\cite{zhou2021domaina, wang2020heterogeneous, xu2021fourier, lu2022fixed}. FIXED~\cite{lu2022fixed} identifies two limitations of applying mixup to DG. The first limitation concerns the challenge of distinguishing between domain and class information, resulting in performance degradation due to entangled domain-class knowledge. The second limitation is that mixup can produce noisy data, particularly when data points from different classes are closely positioned. Inspired by this research, we restrict the application of Uncertainty-adaptive Domain Mix (UDMix) exclusively to images of identical classes through SampleMatch. Then, we implement Prototype Merging Loss (PML), which reduces intra-class distance across varying domains to enhance generalization capabilities, while simultaneously, maintaining a clear separation between data points of different classes for effective discrimination. This approach prevents the entanglement of domain and class knowledge, enabling the effective learning of domain-invariant features that perform class discrimination across datasets from different domains. 

\section{The ProUD Algorithm}
\label{3}
The goal of \textit{Semi-Supervised Domain Generalization} (SSDG) is to learn domain-invariant features from multiple source domains, wherein only one domain is labeled while the other domains are unlabeled. \textit{Semi-supervised learning} and \textit{single domain generalization} are two existing approaches that can be straightforwardly applied to address the SSDG problem.

Semi-supervised learning trains a model using a small amount of labeled data alongside a larger volume of unlabeled data. However, this method assumes all training data originate from a single domain and \textit{lacks a mechanism to utilize domain information}. This impedes the learning of domain-invariant features, thereby hindering generalization.

Another approach to consider is single domain generalization, which aims to achieve generalization using a single labeled source domain. This approach can be applied to the SSDG problem in a way that exclusively uses the labeled domain among the source domains. However, \textit{its lack of mechanism to utilize unlabeled domains} may result in suboptimal performance, particularly in domains with low relevance to the labeled source domain.

To address these limitations of existing approaches in solving SSDG, we propose the Prototype-based Uncertainty-adaptive Domain Generalization (ProUD) algorithm, which is detailed in Algorithm~\ref{alg}. To leverage both domain information and unlabeled domains, our approach incorporates two key strategies: 
\vspace{\baselineskip}

\noindent(1) Learning domain-invariant features via domain-aware prototypes
\begin{itemize}
\item To effectively utilize domain information from training data for the learning of domain-invariant features, we introduce Domain-aware Prototype-based Pseudo-labeling (DaPP) and Prototype Merging Loss (PML).
\item DaPP employs domain-aware prototypes to assign pseudo-labels to samples from unlabeled domains, which generates more accurate pseudo-labels even when the unlabeled domains have significant discrepancies between them.
\item PML guides data from the same class across different domains to cluster around a single class prototype, resulting in both the features and their class prototypes to become domain-invariant.
\end{itemize}

\noindent (2) Uncertainty-adaptive mixing of labeled and unlabeled domains
\begin{itemize}
\item To effectively utilize unlabeled domains, we introduce SampleMatch and Uncertainty-adaptive Domain Mix (UDMix).
\item SampleMatch pairs each pseudo-labeled sample with a labeled sample of the corresponding class to generate a domain-mixed sample, which helps enhancing robustness against noisy pseudo-labels.
\item UDMix modulates the involvement of each pseudo-labeled sample from different domains based on its uncertainty, which enables progressive generalization over multiple unlabeled domains in a robust manner. 


\end{itemize}

In the subsequent sections, we delve into the specifics of our problem and algorithm introduced earlier. Sec.~\ref{problem} is dedicated to defining the problem formulation for SSDG, laying the groundwork for our discussion and experiments. Following this, we provide an in-depth exploration of the ProUD algorithm, detailing the core ideas and algorithmic processes of its two strategies in Sec.~\ref{algorithm1} and \ref{algorithm2}, respectively.

\begin{algorithm} [t]
\begin{algorithmic}
    \State {\bfseries Input:} Pretrained model $f=h \circ g$, labeled source domain dataset $\mathcal{D}^l =\mathcal{D}_0$, unlabeled source domain datasets $\mathcal{D}^u=\{\mathcal{D}_1, \dots, \mathcal{D}_T\}$, and balancing parameter $\alpha$.
    \For{epoch $=1$ to $E$}\Comment{for each epoch} 
        \For{$t=1$ to $T$} \Comment{for each unlabeled source domain}
        \State Apply $\text{DaPP}$ on $\smash{\mathcal{D}_t}$ to generate domain-aware prototypes $\smash{\{C_{t,k}\}_{k=1}^{K}}$.   
        \State Build a pseudo-labeled dataset $\smash{\widetilde{\mathcal{D}}_t}$.
        \State Estimate uncertainty $\epsilon(\vx_t)$ for all $\vx_t \in \smash{\widetilde{\mathcal{D}}_t}$.
        \EndFor
        \State Sample a sequence of mini-batches $\smash{\{\mathcal{B}_s^u\}_{s=1}^S}$ from $\smash{\bigcup_{t=1}^{T}\widetilde{\mathcal{D}}_t}$.
        \For{$s=1$ to $S$}\Comment{for each mini-batch} 
            \State $\mathcal{B}_s^l=\text{SampleMatch}(\mathcal{B}_s^u,\mathcal{D}_0)$
            \State $\mathcal{B}_s^m=\text{UDMix}(\mathcal{B}_s^l, \mathcal{B}_s^u)$
            \State Update $f$ with $\mathcal{L}(h \circ g)=\mathcal{L}_{\text{CE}}(h \circ g;\mathcal{B}_s^m)+\alpha \mathcal{L}_{\text{PML}}(g;\mathcal{B}_s^l\cup\mathcal{B}_s^u)$.
        \EndFor
    \EndFor
\end{algorithmic}
\caption{The ProUD Algorithm}
\label{alg}
\end{algorithm}

\subsection{Semi-Supervised Domain Generalization}
\label{problem}
In this work, we tackle the $K$-way image classification problem for SSDG. The image space $\mathcal{X}$ and label space $\mathcal{Y}$ are assumed to be shared across all domains. For notational simplicity, the notation $\mathcal{D}_t$ denotes both the $t$-th domain and the dataset sampled from it interchangeably. To formulate the SSDG problem, consider a training dataset $\mathcal{D}^\text{train}=\mathcal{D}^l \cup \mathcal{D}^u$, consisting of a single labeled source domain $\mathcal{D}^l$ and a set of unlabeled source domains $\mathcal{D}^u$. Specifically, we have a labeled source domain $\smash{\mathcal{D}^l = \mathcal{D}_0 = \{(\vx_0^{(i)}, y_0^{(i)})\}_{i=1}^{N_0}}$, where $N_0$ denotes the total number of samples in $\mathcal{D}_0$, and $\vx_0^{(i)}$ and $y_0^{(i)}$ indicate input data and its corresponding label of the $i$-th sample. On the other hand, a set of unlabeled source domains $\mathcal{D}^u= \{\mathcal{D}_t\}_{t=1}^{T}$ does not contain any label information (\textit{i.e.,} $\mathcal{D}_t= \{\vx_t^{(i)}\}_{i=1}^{N_t}$). In this setting, we have a model $f(\vx)=(h \circ g)(\vx)$, where $g: \sX \rightarrow \sR^d$ and $h: \sR^d \rightarrow \sR$ represent the feature extractor and the classifier, respectively. $d$ denotes the feature dimension. The goal of SSDG is to train the model $f$ with $\mathcal{D}^\text{train}$ to generalize well on a test dataset from unseen domain $\mathcal{D}^\text{test}$, where $\mathcal{D}^\text{train} \cap \mathcal{D}^\text{test} = \emptyset$.

\begin{figure}[t]
\begin{center}
\includegraphics[width=\linewidth]{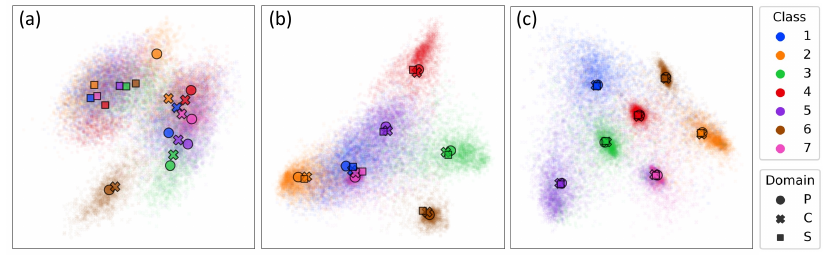}

\end{center}
   \caption{t-SNE visualizations of the learned representations of both samples and domain-aware prototypes using the PACS dataset, in the case where P is the labeled source domain, and C and S are the unlabeled source domains. Different colors and shapes represent distinct classes and domains, respectively. 
   (a), (b), and (c) are produced right after the DaPP at epochs 1, 40, and 80, respectively.}
\label{fig:tsne}
\end{figure}

\subsection{Learning Domain-Invariant Features via Domain-Aware Prototypes} 
\label{algorithm1}
\subsubsection{Core Idea}
To effectively utilize the domain information of training data for learning domain-invariant features, we propose Domain-aware Prototype-based Pseudo-labeling (DaPP) and Prototype Merging Loss (PML). First, we introduce DaPP to assign pseudo-labels to samples from unlabeled domains $\mathcal{D}^u$. The core idea of DaPP lies in its use of \textit{domain-aware prototypes}, generating distinct prototypes for each domain. This approach is crucial, particularly when there are significant discrepancies between domains, as features can be separated more by domains rather than by classes. 

For instance, in Fig.~\ref{fig:tsne}(a), the class 1 prototype from domain S, represented by a blue square, is significantly distant from the class 1 prototypes of domains P and C, represented by a blue circle and cross, respectively. Meanwhile, it is closely clustered with other class prototypes within domain S, depicted by squares of various colors. In such scenarios, constructing class prototypes without distinguishing between domains could result in the prototypes not being closely situated with their corresponding features, thereby generating inaccurate pseduo-labels. 

Starting with Fig.~\ref{fig:tsne}(a), our ultimate goal is to learn domain-invariant representations, where data from the same class across different domains cluster around a single class prototype. To this end, we employ PML based on contrastive learning, which encourages prototypes of the same class to cluster together while distancing them from others. Fig.~\ref{fig:tsne}(b) and (c) show that, as the number of training epochs increases, PML effectively merges the prototypes from different domains into a single point, which leads both the features and their class prototypes to become domain-invariant.

\subsubsection{Algorithm}

Given labeled source domain $\mathcal{D}^l =\mathcal{D}_0$ and unlabeled source domains $\mathcal{D}^u=\{\mathcal{D}_1, \dots, \mathcal{D}_T\}$, we first pretrain the model $f$ using $\mathcal{D}^l$, as detailed in the supplementary material. In the beginning of every training epoch, we apply DaPP to generate a domain-aware class prototype $C_{t,k}$, which indicates the centroid of features for samples in class $k$ for domain $\mathcal{D}_t$,
\begin{equation}
\label{eq:1}
C_{t,k} = \frac{\sum_{\vx_t \in \mathcal{D}_t}\delta_k(f(\vx_t))\bar{g}(\vx_t)}{\sum_{\vx_t \in \mathcal{D}_t}\delta_k(f(\vx_t))},
\end{equation}
where $\delta_k(\cdot)$ denotes the $k$-th element of a softmax output, and $\bar{g}(\cdot)=g(\cdot)/\lVert g(\cdot)\rVert$. Then, each unlabeled sample $\vx_t$ is pseudo-labeled as
\begin{equation}
\label{eq:pl}
\hat{y}(\vx_t)=\argmin_k \dist(\bar{g}(\vx_t),C_{t,k}),
\end{equation}
where $\dist$ is the cosine distance between the feature of sample $\vx_t$ and the prototypes of its domain $\mathcal{D}_t$. Inspired by \cite{liang2020we}, we reconstruct the prototypes based on the new
pseudo-labels assigned in Eq.~\ref{eq:pl} as
\begin{equation}
C_{t,k} = \frac{\sum_{\vx_d \in \mathcal{D}_t}\mathds{1}(\hat{y}(\vx_t)=k)\bar{g}(\vx_t)}{\sum_{\vx_t \in \mathcal{D}_t}\mathds{1}(\hat{y}(\vx_t)=k)}.
\end{equation}
Based on the new prototypes, $C_{t,1},\cdots,C_{t,K}$ for $\mathcal{D}_t$ ($t>0$), we use Eq.~\ref{eq:pl} to construct a pseudo-labeled dataset $\widetilde{\mathcal{D}}_t=\{(\vx_t^{(i)},\hat{y}_t^{(i)}\}_{i=1}^{N_t}$. In this step, an ensemble of predictions from different augmentations can be used for robustness as described in the supplementary material. Next, we train the feature extractor $g$, based on PML, defined as
\begin{equation}
    \mathcal{L}_\text{PML}(g;\mathcal{B}^l\cup\mathcal{B}^u) = -\sum_{(\vx,y) \in \mathcal{B}^l\cup\mathcal{B}^u}\log\frac{\exp\big(-\dist(\bar{g}(\vx),\overline{C}_y)\big)}{\sum_{k}\exp\big(-\dist(\bar{g}(\vx),\overline{C}_k)\big)},
\label{eq:pml}
\end{equation}
where $\mathcal{B}^l$ and $\mathcal{B}^u$ represent a batch of labeled samples and a batch of unlabeled samples, respectively, and $\overline{C}_k=\langle C_{t,k} \rangle_t$. The notation $\langle \; \cdot \; \rangle_t$ denotes an average over all values of $t$. $\overline{C}_k$ plays a pivotal role as an anchor point, attracting the features of all samples that belong to class $k$, regardless of the domains they originate from. Consequently, PML facilitates the merging of prototypes from different domains into a single point, and the feature extractor $g$ is guided to learn domain-invariant representations as illustrated in Fig.~\ref{fig:tsne}(c).

\subsection{Uncertainty-Adaptive Mixing of Labeled and Unlabeled Domains}
\label{algorithm2}

\begin{wrapfigure}{h}{0.5\textwidth}
\vspace{-30pt}
\begin{center}
\includegraphics[width=0.5\textwidth]{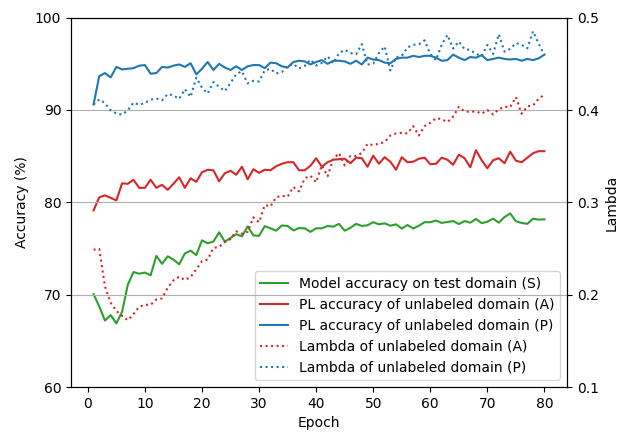}
\end{center}
\vspace{-20pt}
\caption{Training curves of ProUD on the PACS dataset, with labeled source domain C, unlabeled source domains A and P, and the test domain S. Lambda and PL accuracy respectively represent the average value of the mixing ratio $\lambda$ and the accuracy of pseudo-labels across all samples within each of the unlabeled domains.}
\label{fig:training}
\vspace{-10pt}
\end{wrapfigure}

\subsubsection{Core Idea}
A significant challenge in utilizing unlabeled domain data is ensuring model robustness against noisy pseudo-labels. To address this issue, we introduce SampleMatch, a method that pairs each pseudo-labeled sample from unlabeled domains with a corresponding class sample from the labeled domain. These pairs are then mixed via Uncertainty-adaptive Domain Mix (UDMix) to generate domain-mixed images for training. The core idea behind UDMix is to limit the influence of a pseudo-labeled sample if there is high uncertainty regarding its correct labeling. To achieve this, we adjust the mixing ratio between the pair of labeled and pseudo-labeled samples based on the uncertainty of the pseudo-labeled sample. Specifically, when the uncertainty is high--indicating a higher likelihood of the pseudo-label being incorrect--the mixing ratio $\lambda$ of the pseudo-labeled image is reduced, thereby increasing the proportion of the labeled image. This approach prevents potential performance degradation due to training with incorrect pseudo-labels, as illustrated by the training examples in Fig.~\ref{fig:main}.

The training curves of ProUD in Fig~\ref{fig:training} show how the mixing ratio $\lambda$ is adjusted for each domain with UDMix. Initially, the pseudo-label accuracy in the unlabeled domain P exceeds that of domain A by more than ten percentage points. This disparity stems from the higher relevance between domain P and the labeled domain C. Given the high uncertainty inherent in noisy labels, UDMix modulates the mixing ratio $\lambda$ to be higher for domain P than domain A. Consequently, samples with correct pseudo-labels are more actively engaged in the training process than those with incorrect pseudo-labels. As the model generalizes better with correct labels, it leads to an increase in samples with accurate pseudo-labels, which, in turn, encourages more pseudo-labeled data to participate in training, further enhancing model generalization. This positive feedback loop is evident in the curves depicting gradually improving pseudo-label accuracy and mixing ratio $\lambda$ as the number of training epochs increases. The enhanced generalization capacity of the model is further confirmed by the progressively improved accuracy on the test domain S. 

\subsubsection{Algorithm}
Given a sample $(\vx_t,\hat{y}_t)$ from pseudo-labeled dataset $\widetilde{\mathcal{D}}_{t}$, we first estimate the entropy-based uncertainty, defined as
\begin{equation}
\epsilon(\vx_t)=-\sum_{k}\delta_k(-\dist(\bar{g}(\vx_t),C_{t,k})/\tau_\epsilon)\log \delta_k(-\dist(\bar{g}(\vx_t),C_{t,k})/\tau_\epsilon),
\label{eq:uncertainty}
\end{equation}
where $\epsilon(\vx_t)$ is the uncertainty estimate of sample $\vx_t$ and $\tau_\epsilon$ is a temperature parameter. Then, we randomly sample $\vx_0$ with $y_0=\hat{y}_t$ from the labeled domain $\mathcal{D}_0$. For a batch of pseudo-labeled samples $\smash{\mathcal{B}^u \subset \bigcup_{t=1}^{T}\widetilde{\mathcal{D}}_t}$, we can apply the same sampling process for each sample to find a batch of the corresponding labeled samples $\mathcal{B}^l=\text{SampleMatch}(\mathcal{B}^u,\mathcal{D}_0)$. Finally, we mix each pair of the samples $(\vx^u, \vx^l)$ from $\mathcal{B}^u$ and $\mathcal{B}^l$ to get a batch of samples $\mathcal{B}^m=\text{UDMix}(\mathcal{B}^l, \mathcal{B}^u)$ by the following equation:
\begin{equation}
\vx^m = \lambda \vx^u + (1-\lambda) \vx^l,
\label{eq:UDMix}
\end{equation}
where $\vx^m$ is a domain-mixed sample, and $\lambda$ is the mixing ratio. $\lambda$ is primarily determined by the uncertainty estimate $\epsilon$ of the pseudo-labeled sample $\vx^u$ through $\lambda_\epsilon = \exp (-\epsilon/\tau_{\lambda})/(1+\exp(-\epsilon/\tau_{\lambda}))$, where $\tau_{\lambda}$ is a temperature parameter. Note that this equation implies that the uncertainty of labeled samples is assumed to be zero, which is the minimum value defined by Eq.~\ref{eq:uncertainty}. However, due to the inherent noisiness of the real-world data, it is not feasible for the uncertainty of unlabeled domains to be exactly zero. Thus, we introduce a threshold $\lambda^*$, and if $\lambda_\epsilon>\lambda^*$, we assign a random mixing ratio to the sample to promote more diverse augmentation, formulated as
\begin{equation}
  \lambda = 
  \begin{cases} 
    \lambda_\ru \sim U(0,1), & \text{if } \lambda_\epsilon > \lambda^*, \\
    \lambda_\epsilon, &  \text{otherwise}, 
  \end{cases}
\end{equation}
where $\lambda_\ru$ represents a random value sampled from $U(0,1)$, a uniform random distribution between $0$ and $1$. This way, we can address the case where a larger portion of $\vx^u$ is included in $\vx^m$ than $\vx^l$ for samples with low uncertainty. Finally, we update the model $f$ with the conventional cross entropy loss for domain-mixed samples, $\mathcal{L}_{\text{CE}}(h \circ g;\mathcal{B}^m)$, where $\mathcal{B}^m$ represents a batch of domain-mixed samples. $\mathcal{L}_{\text{CE}}$ is coupled with mixup in the feature level to enhance sample diversity for classification~\cite{zhang2017MixUp}.

\section{Experiments}
\label{4}
We evaluate the effectiveness of our proposed method, ProUD, by comparing against strong baselines on three datasets: PACS~\cite{li2017deeper}, Digits-DG~\cite{lecun1998gradient, ganin2016domain, netzer2011reading, roy2018effects}, and Office-Home~\cite{venkateswara2017deep}. 
Following the experimental setup from ~\cite{lin2024semi}, we design our experiments to address SSDG, a representative case of \textit{data inequality problem across domains in which only a sinlge domain is labeled while the rest are unlabeled}.

\subsubsection{Datasets}
All three datasets consist of four distinct domains.
PACS includes Photo (P), Art Painting (A), Cartoon (C), and Sketch (S) with 7 object categories. 
Digits-DG contains images from MNIST (Mn), MNIST-m (Mm), SVHN (Sv), and SYN-D (Sy) with 10 digits from 0 to 9. 
Office-Home consists of 65 object categories from Art (Ar), Clipart (Cl), Product (Pr), and Real-World (Rw).
We split each source domain dataset ($\mathcal{D}^\text{train}$) into training and validation sets with ratio 9:1 for PACS, 8:2 for Digits-DG, and approximately 9:1 for Office-Home. 

\subsubsection{Experimental Settings}
For each dataset, we conduct experiments for all 12 possible domain combinations across four distinct domains, including one labeled and two unlabeled source domains ($\mathcal{D}^\text{train}$), as well as one unseen test domain ($\mathcal{D}^\text{test}$). 
All experiments are repeated three times with different random seeds (2022/2023/2024), and we report average accuracy obtained from these repetitions.
In each experiment, we calculate the average accuracy over the last five epochs.
We train the model using the SGD optimizer with mini-batches of 128 samples for all datasets.
Further implementation details are provided in the supplementary material.

\subsection{Effectiveness of ProUD}
\label{4.1}
\subsubsection{Baseline Methods}
To evaluate the effectiveness of ProUD, we benchmark it against a comprehensive set of baselines, using the experiment results reported in~\cite{lin2024semi}. The baseline methods include various methods from the literature on semi-supervised learning (MeanTeacher~\cite{tarvainen2017mean}, MixMatch~\cite{berthelot2019mixmatch}, and FixMatch~\cite{sohn2020fixmatch}), single domain generalization (RSC~\cite{huang2020self}, L2D~\cite{wang2021learning}, and DGvGS~\cite{mansilla2021domain}), and semi-supervised domain generalization (EID~\cite{lin2024semi}). 

\begin{table}[t!]
\caption{Comparison of the performance between ours and various baseline methods on the PACS dataset. 
The bullet $^\bullet$ and asterisk $^*$ indicate labeled source domain and unseen test domain, respectively. All other domains are unlabeled source domains.
Avg. and Std. respectively stand for average and standard deviation across 12 domain combinations.
The best results are in bold. This format persists in subsequent tables.}

\label{table:1}
\makebox[\textwidth]{
\begin{adjustbox}{width=1.0\textwidth}
\setlength\extrarowheight{2pt}
\begin{tabular}{c|c|ccc|ccc|ccc|ccc|c|c}
\hline
\multirow{2}{*}{Type} & \multirow{2}{*}{Method} & & P$^\bullet$ &  & & A$^\bullet$ & & & C$^\bullet$ &  &  & S$^\bullet$ &  & \multirow{2}{*}{Avg.} & \multirow{2}{*}{Std.} \\ \cline{3-14}
&  & A$^*$ & C$^*$ & S$^*$ & P$^*$ & C$^*$ & S$^*$ & P$^*$ & A$^*$ & S$^*$ & P$^*$ & A$^*$ & C$^*$ &  & \\\hline
 \multirow{3}{*}{\begin{tabular}[c]{@{}c}Semi-Supervised \\ Learning\end{tabular}} & MeanTeacher & 54.7 & 36.3 & 33.1 & 91.9 & 65.6 & 38.1 & 80.3 & 60.6 & 58.6 & 38.1 & 33.4 & 54.7 & 53.8 & 18.3 \\
 & MixMatch & 35.7 & 16.2 & 24.5 & 87.3 & 62.7 & 47.6 & 43.1 & 47.9 & 50.7 & 26.1 & 46.9 & 52.2 & 45.1 & 18.0 \\
 & FixMatch & 66.8 & 34.9 & 25.9 & \textbf{96.6} & 72.9 & 67.1 & \textbf{91.7} & 76.5 & 69.5 & 36.3 & 35.2 & 56.0 & 60.8 & 22.3 \\ \hline
 \multirow{3}{*}{\begin{tabular}[c]{@{}c}Single DG\end{tabular}} & RSC & 66.6 & 27.6 & 38.6 & 93.7 & 68.0 & 65.7 & 83.5 & 69.2 & 76.6 & 47.5 & 43.0 & 65.2 & 62.1 & 18.5 \\
 & L2D & 65.2 & 30.7 & 35.4 & 96.1 & 65.7 & 58.0 & 87.3 & 73.5 & 67.9 & 48.2 & 45.9 & 61.8 & 61.3 & 18.6 \\
 & DGvGS & 54.2 & 16.6 & 28.5 & 93.8 & 54.7 & 39.7 & 80.3 & 59.5 & 56.7 & 14.3 & 16.2 & 17.2 & 44.3 & 25.5 \\ \hline
\multirow{2}{*}{\begin{tabular}[c]{@{}c} Semi-Supervised\\ DG\end{tabular}} & EID & \textbf{75.5} & \textbf{71.0} & 64.0 & 94.9 & 71.8 & 67.2 & 84.6 & 77.4 & 72.2 & \textbf{67.2} & 66.9 & \textbf{72.8} & 73.8 & 8.3 \\ \cline{2-16} 
 & \textbf{ProUD (Ours)} & 73.8 & 63.6 & \textbf{74.1} & 91.1 & \textbf{75.4} & \textbf{76.6} & 86.9 & \textbf{78.9} & \textbf{78.1} & 63.0 & \textbf{68.9} & 70.4 & \textbf{75.1} & \textbf{8.0} \\ \hline
\end{tabular}
\end{adjustbox}}
\end{table}

\begin{table}[t!]
\caption{Comparison of the performance between ours and various baseline methods on the Digits-DG dataset.}

\label{table:2}
\makebox[\textwidth]{
\begin{adjustbox}{width=1.0\textwidth}
\setlength\extrarowheight{2pt}
\begin{tabular}{c|c|ccc|ccc|ccc|ccc|c|c}
\hline
\multirow{2}{*}{Type} & \multirow{2}{*}{Method} & & Mn$^\bullet$ &  & & Mm$^\bullet$ & & & Sv$^\bullet$ &  &  & Sy$^\bullet$ &  & \multirow{2}{*}{Avg.} & \multirow{2}{*}{Std.} \\ \cline{3-14}
&  & Mm$^*$ & Sv$^*$ & Sy$^*$ & Mn$^*$ & Sv$^*$ & Sy$^*$ & Mn$^*$ & Mm$^*$ & Sy$^*$ & Mn$^*$ & Mm$^*$ & Sv$^*$ &  & \\\hline
\multirow{3}{*}{\begin{tabular}[c]{@{}c}Semi-Supervised \\ Learning\end{tabular}} & MeanTeacher & 23.9 & 13.8 & 26.4 & 82.1 & 19.3 & 32.4 & 43.6 & 17.5 & 59.2 & 56.3 & 22.2 & 38.4 & 36.3 & 19.8 \\
 & MixMatch & 33.0 & 18.1 & 30.3 & 93.6 & 26.7 & 45.4 & 59.3 & 27.9 & 76.7 & 67.7 & 36.9 & 51.5 & 47.3 & 22.1 \\
 & FixMatch & 29.9 & 10.6 & 23.9 & 90.8 & 32.5 & 48.2 & 57.5 & 40.0 & 70.9 & 74.0 & 51.9 & 61.3 & 49.3 & 22.2 \\ \hline
 \multirow{3}{*}{\begin{tabular}[c]{@{}c}Single DG\end{tabular}} & RSC & 42.8 & 19.3 & 45.0 & 93.6 & 11.7 & 12.0 & 70.5 & 46.1 & \textbf{95.5} & 81.4 & 42.4 & 78.8 & 53.3 & 29.0 \\
 & L2D & 57.2 & 28.2 & 53.1 & 97.1 & 12.5 & 25.1 & 72.4 & 52.8 & 94.2 & 80.2 & 45.7 & \textbf{80.0} & 58.2 & 26.2 \\
 & DGvGS & 13.3 & 12.2 & 19.5 & 88.9 & 11.2 & 16.5 & 57.7 & 24.2 & 88.5 & 68.4 & 25.9 & 67.1 & 41.1 & 29.3 \\ \hline
\multirow{2}{*}{\begin{tabular}[c]{@{}c}Semi-Supervised \\ DG\end{tabular}} & EID & 51.6 & 37.3 & 53.3 & 97.1 & 58.6 & 69.1 & \textbf{87.7} & 60.9 & 87.5 & \textbf{92.4} & 64.2 & 70.9 & 69.2 & 17.8 \\ \cline{2-16} 
& \textbf{ProUD (Ours)} & \textbf{70.3} & \textbf{57.8} & \textbf{65.8} & \textbf{97.6} & \textbf{63.4} & \textbf{70.5} & 80.7 & \textbf{61.5} & 83.4 & 92.2 & \textbf{65.5} & 74.4 & \textbf{73.6} & \textbf{12.0} \\ \hline
\end{tabular}
\end{adjustbox}}
\end{table}

\begin{table}[t!]
\caption{Comparison of the performance between ours and various baseline methods on the Office-Home dataset.}

\label{table:3}
\makebox[\textwidth]{
\begin{adjustbox}{width=1.0\textwidth}
\setlength\extrarowheight{2pt}
\begin{tabular}{c|c|ccc|ccc|ccc|ccc|c|c}
\hline
\multirow{2}{*}{Type} & \multirow{2}{*}{Method} & & Ar$^\bullet$ &  & & Cl$^\bullet$ & & & Pr$^\bullet$ &  &  & Rw$^\bullet$ &  & \multirow{2}{*}{Avg.} & \multirow{2}{*}{Std.} \\ \cline{3-14}
&  & Cl$^*$ & Pr$^*$ & Rw$^*$ & Ar$^*$ & Pr$^*$ & Rw$^*$ & Ar$^*$ & Cl$^*$ & Rw$^*$ & Ar$^*$ & Cl$^*$ & Pr$^*$ &  & \\\hline
\multirow{3}{*}{\begin{tabular}[c]{@{}c}Semi-Supervised \\ Learning\end{tabular}} & MeanTeacher & 35.1 & 50.5 & 60.8 & 39.1 & 51.4 & 54.0 & 35.8 & 34.5 & 62.0 & 54.4 & 43.4 & 72.2 & 49.4 & 11.6 \\
 & MixMatch & 40.0 & 51.8 & 62.4 & 43.2 & 57.6 & 58.9 & 42.0 & 38.5 & 63.6 & 55.5 & 43.7 & 72.4 & 52.5 & 10.5 \\
 & FixMatch & 41.4 & 55.3 & 64.4 & 44.4 & 57.8 & 57.5 & 44.0 & 42.2 & 65.8 & \textbf{57.2} & 45.0 & \textbf{73.7} & 54.1 & 10.2 \\ \hline
\multirow{3}{*}{\begin{tabular}[c]{@{}c}Single DG\end{tabular}} & RSC & 39.1 & 49.8 & 61.1 & 36.9 & 53.0 & 53.7 & 35.9 & 38.8 & 61.2 & 53.3 & 45.6 & 72.2 & 50.1 & 10.8 \\
 & L2D & 39.6 & 44.8 & 57.5 & 42.2 & 52.6 & 55.7 & 38.5 & 43.0 & 62.3 & 55.0 & 48.3 & 69.3 & 50.7 & 9.2 \\
 & DGvGS & 33.4 & 42.9 & 55.4 & 32.6 & 45.0 & 47.0 & 29.8 & 33.2 & 55.0 & 50.8 & 37.9 & 68.0 & 44.3 & 11.1 \\ \hline

\multirow{2}{*}{\begin{tabular}[c]{@{}c} Semi-Supervised \\ DG \end{tabular}} & EID & 48.3 & \textbf{59.1} & \textbf{66.5} & \textbf{47.5} & \textbf{60.4} & \textbf{61.3} & 46.1 & 47.3 & \textbf{66.0} & 53.3 & 48.8 & 69.0 & 56.1 & 8.2 \\ \cline{2-16} 
& \textbf{ProUD (Ours)}& \textbf{48.5} & 57.4 & 65.0 & 46.0 & 57.7 & 59.9 & \textbf{47.0} & \textbf{49.4} & 65.3 & 55.1 & \textbf{52.5} & 71.9 & \textbf{56.3} & \textbf{7.8} \\ \hline
\end{tabular}
\end{adjustbox}}
\end{table}

\subsubsection{Results}
We present an evaluation of our ProUD algorithm against the baseline methods on the three datasets (PACS, Digits-DG, and Office-Home datasets). The results for each dataset are detailed in Tabs.~\ref{table:1},~\ref{table:2}, and~\ref{table:3}, respectively. The average accuracy and standard deviation across 12 distinct domain combinations are denoted as Avg. and Std., respectively. As illustrated in these tables, ProUD consistently outperforms all baseline models in terms of accuracy across all datasets. It is particularly noteworthy that most existing methods show a lack of robustness across different domain combinations. For example, in Tab.~\ref{table:2}, while RSC achieves the best performance for the Sv$^\bullet$/Sy$^*$ combination, it exhibits the lowest performance among all baselines for the Mm$^\bullet$/Sy$^*$ combination. In contrast, ProUD demonstrates its robustness in Tabs.~\ref{table:1},~\ref{table:2}, and~\ref{table:3}, consistently achieving the lowest standard deviation over different domain combinations across all datasets.

\subsubsection{Comparison with State-of-the-Art in SSDG}
All three experiments demonstrate that our method outperforms EID~\cite{lin2024semi}, the current state-of-the-art in SSDG, in terms of average performance and robustness across different domain combinations. Beyond outperforming EID, another crucial advantage of our method is its \textit{scalability}. In the case of EID, it requires as many separate domain adaptation models and training processes as there are unlabeled source domains. This requirement imposes a significant limitation on its practical application to SSDG problems with a large number of source domains. In contrast, ProUD’s progressive generalization algorithm allows for handling multiple unlabeled source domains simultaneously with a single model in a scalable manner through the uncertainty-adaptive mixing of labeled and unlabeled domains.

\begin{table}[t]
\caption{Ablation study of UDMix and PML on the PACS dataset. Diff. denotes the difference in comparison to our model, $\proud$, incorporating both UDMix and PML.}
\label{table:4}
\begin{adjustbox}{width=\textwidth}
\setlength\extrarowheight{2pt}
\begin{tabular}{c|ccc|ccc|ccc|ccc|c|c}
\hline
\multirow{2}{*}{Method} & & P$^\bullet$ &  & & A$^\bullet$ & & & C$^\bullet$ &  &  & S$^\bullet$ &  & \multirow{2}{*}{Avg.} & \multirow{2}{*}{Std.} \\ \cline{2-13}
  & A$^*$ & C$^*$ & S$^*$ & P$^*$ & C$^*$ & S$^*$ & P$^*$ & A$^*$ & S$^*$ & P$^*$ & A$^*$ & C$^*$ &  & \\ \hline
  ProUD (Ours) & 73.8 & 63.6 & 74.1 & 91.1 & 75.4 & 76.6 & 86.9 & 78.9 & 78.1 & 63.0 & 68.9 & 70.4 & 75.1 & 8.0\\ \hline
w/o UDMix & 68.9 & 53.3 & 63.3 & 90.9 & 72.3 & 72.8 & 82.0 & 79.1 & 76.7 & 51.4 & 60.2 & 70.0 & 70.1 & 11.1 \\
Diff. & -4.9 & -10.2 & -11.4 & -0.4 & -3.2 & -3.3 & -5.0 & 0.0 & -1.1 & -11.3 & -9.3 & +0.1 & -5.0 & +3.1 \\ \hline
w/o PML & 70.5 & 59.6 & 62.5 & 91.5 & 73.3 & 74.0 & 87.1 & 78.9 & 79.0 & 62.4 & 68.6 & 70.2 & 73.1 & 9.3 \\
Diff. & -3.3 & -3.9 & -12.2 & +0.2 & -2.2 & -2.1 & +0.1 & -0.2 & +1.2 & -0.3 & -0.9 & +0.3 & -2.0 & +1.3 \\ \hline
\end{tabular}
\end{adjustbox}
\end{table}

\subsection{Further Analysis}

\subsubsection{Ablation Study of UDMix}
To validate the effectiveness of UDMix, we perform an ablation study using the PACS dataset, removing the strategy that determines the mixing coefficient $\lambda$ based on the uncertainty of pseudo-labels. Instead, $\lambda$ is sampled from a uniform distribution ranging from $0$ to $1$ throughout the training. As shown in Tab.~\ref{table:4}, excluding UDMix results in significantly decreased average performance and increased standard deviation. This underscores its crucial contribution to the model's performance and robustness in ProUD.

\begin{figure}[t!]
\begin{center}
\includegraphics[width=\linewidth]{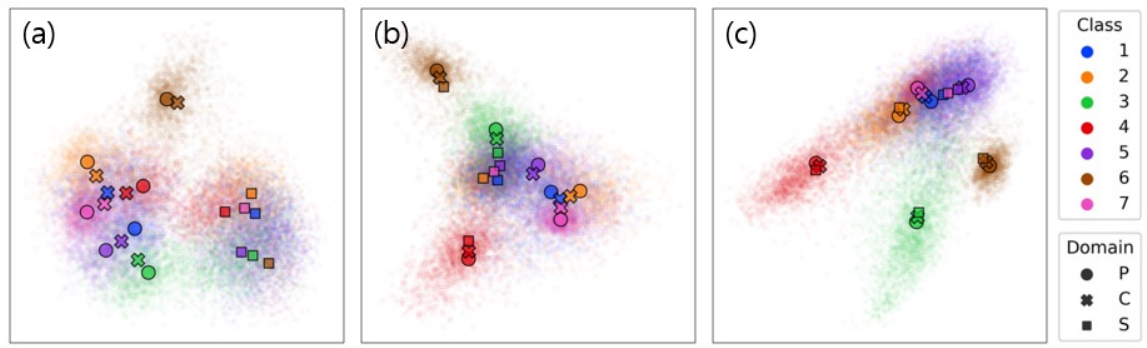}

\end{center}
   \caption{t-SNE visualizations of the learned representations of both samples and domain-aware prototypes \textit{without PML}. P is the labeled source domain, and C and S are the unlabeled source domains from the PACS dataset. Different colors and shapes represent distinct classes and domains, respectively. (a), (b), and (c) are produced right after the DaPP at epochs 1, 40, and 80, respectively.}
\label{fig:tsne2}
\end{figure}

\subsubsection{Ablation Study of PML}
We conduct an ablation study to assess the impact of PML within ProUD, by comparing the performance of models trained both with and without this loss term. The results are shown in Tab.~\ref{table:4}. The exclusion of the loss term leads to a decrease in average accuracy and an increase in standard deviation. For a more in-depth analysis, we present a t-SNE visualization of learned representations \textit{without PML} in Fig.~\ref{fig:tsne2}. Contrary to the well-separated clusters observed \textit{with PML} in Fig.~\ref{fig:tsne}, the samples are only loosely clustered around their respective prototypes, resulting in more ambiguous cluster boundaries. Even after 80 epochs, several prototypes remain intermingled, which leads to a degradation in both performance and robustness.

\section{Conclusion}
In this paper, we address a representative case of the data inequality problem across domains, termed Semi-Supervised Domain Generalization (SSDG), where only one domain is labeled, leaving the others unlabeled. Such a setting mirrors real-world situations, especially when obtaining labeled data from certain domains is considerably more difficult than from others. Moreover, such data inequality not only presents practical challenges but also raises ethical concerns in the design and deployment of machine learning models. To overcome this issue of data inequality, we propose the ProUD alogrithm, which can effectively learn domain-invariant features via domain-aware prototypes along with progressive generalization via uncertainty-adaptive mixing of labeled and unlabeled domains. Extensive experiments demonstrate that ProUD outperforms all baseline methods in terms of both model performance and robustness. The primary focus of this paper has been to validate the general applicability of our method through experiments using standard benchmark datasets. We hope that future research will expand the application of this method to a broad spectrum of areas vulnerable to data inequality across domains, such as biomedical imaging and autonomous driving.

\newpage


%
%
\bibliographystyle{splncs04}
\bibliography{main}

\clearpage
\newpage



\section*{Supplementary material}

# Supplementary Material: Overcoming Data Inequality across Domains with Semi-Supervised Domain Generalization

Jinha Park[*1], Wonguk Cho[*2], and Taesup Kim[2]

[1] Department of Electrical and Computer Engineering, Seoul National University
jhpark410@snu.ac.kr
[2] Graduate School of Data Science, Seoul National University
{wongukcho, taesup.kim}@snu.ac.kr

## 1 Implementation Details

*Experimental Settings* We conduct experiments on three datasets: PACS, Digits-DG, and Office-Home, training the model for 80, 800, and 60 epochs each. For optimization, we use the SGD optimizer, with a momentum of 0.9 and weight decays of 0.01, 0.005, and 0.0001 corresponding to each dataset. The learning rate is set to 0.001 for PACS and Office-Home, and 0.01 for Digits-DG.

For PACS and Office-Home, we use simple augmentation techniques including random translation, horizontal flipping, color jittering, and grayscaling. The same augmentations are applied to the Digits-DG dataset, except for horizontal flipping. In DaPP, each dataset utilizes three different augmentations for ensemble learning ($r$), while the temperature parameter ($\tau_\epsilon$) is 0.03, 0.1, and 0.02 for PACS, Digits-DG, and Office-Home, respectively. For UDMix, the threshold of $\lambda$ ($\lambda^*$) is set at 0.35, and the temperature parameter ($\tau_\lambda$) is 0.5, consistent across all datasets. The balancing parameter in the loss function ($\alpha$) is set to 0.5. We set the mixup hyperparameter to 2.0 for PACS and Office-Home, and to 0.4 for Digits-DG.

*Network Architectures* The network of our model is composed of a feature extractor and a classifier. We utilize ResNet-18 [3] as the feature extractor for both PACS and Office-Home. For Digits-DG, we use a different backbone consisting of four convolutional layers. ResNet-18 is initialized with the weights pretrained on ImageNet [1]. The classifier consists of a single fully connected layer with weight normalization for all three datasets.

## 2 Pretraining

In this section, we explain the details of pretraining of the ProUD algorithm. We first pretrain $f$ with cross-entropy loss, only using labeled data from $\mathcal{D}^l = \mathcal{D}_0$ as an initializer for joint training with the unlabeled domains $\mathcal{D}^u$. To prevent

---
[*]Equal contribution

overfitting to $\mathcal{D}^l$, we introduce a noise mixing augmentation technique named NoiseMix. The core idea of this technique is to mix a given sample $\boldsymbol{x}$ with a randomly initialized convolutional layer $\Phi$, such that $\boldsymbol{x} \leftarrow \lambda\Phi(\boldsymbol{x}) + (1-\lambda)\boldsymbol{x}$, where $\lambda$ is a mixing ratio. The advantage of employing a convolutional layer is that it allows the generation of a noised image without significantly distorting the semantic information of the original image, due to its translational invariance. We apply NoiseMix repeatedly for $P$ steps using a sequence of different random convolutional layers $\Phi_1, \ldots, \Phi_P$.

## 3   Convolution-based Noise Mixing Augmentation

In this section, we explain the details of NoiseMix. We first generate a sequence of convolution layers $(\Phi_p)_{p=1}^P$, where $\Phi_p$ consists of a randomly initialized convolutional layer and an activation function. The random initialization includes Kaiming Uniform parameter initialization [2], as well as the random choice of kernel size $k$ and dilation $d$ from uniform distributions. The zero-padding method is used to preserve the size of the image. The hyperbolic tangent function is used as an activation function, followed by normalization.

In Algorithm 1, the pseudocode of NoiseMix is presented. Given a batch of samples $\mathcal{B}_{\text{org}}$, we obtain a batch of noise augmented samples $\mathcal{B}_{\text{aug}}$. Then, we pretrain the model $f$ by concatenating $\mathcal{B}_{\text{org}}$ and $\mathcal{B}_{\text{aug}}$. Besides, we employ NoiseMix in the process of UDMix. For given $\mathcal{B}_{\text{org}}^l$ and $\mathcal{B}_{\text{org}}^u$, we first generate batches of noise-mixed samples $\mathcal{B}_{\text{aug}}^l$ and $\mathcal{B}_{\text{aug}}^u$. We then apply UDMix to either $\{(\mathcal{B}_{\text{org}}^l, \mathcal{B}_{\text{org}}^u), (\mathcal{B}_{\text{aug}}^l, \mathcal{B}_{\text{aug}}^u)\}$ or $\{(\mathcal{B}_{\text{org}}^l, \mathcal{B}_{\text{aug}}^u), (\mathcal{B}_{\text{aug}}^l, \mathcal{B}_{\text{org}}^u)\}$ with probability of 0.5.

Throughout this work, we use the same setting of NoiseMix for all the experiments on the three datasets: sequence length $P = 3$, min kernel size $k_{\min} = 1$, max kernel size $k_{\max} = 15$, min dilation $d_{\min} = 1$, max dilation $d_{\max} = 2$, min mixing ratio $\lambda_{\min} = 0.1$, max mixing ratio $\lambda_{\max} = 0.3$, inversion probability $p = 0.2$, and noise strength $\sigma = 0.01$.

---

**Algorithm 1** NoiseMix

**Input:** Sample $\boldsymbol{x}_{\text{org}}$, a sequence of randomly initialized convolution layers $\Phi_1, \ldots, \Phi_P$, and a sequence of randomly sampled mixing ratio $\lambda_1, \ldots, \lambda_P$.
$\boldsymbol{x} = \text{RandomInvert}(\boldsymbol{x}_{\text{org}})$
**for** $p = 1$ to $P$ **do**
    Sample a Gaussian noise $\mathbf{n} \sim \mathcal{N}(\mathbf{0}, \sigma \boldsymbol{I})$
    $\tilde{\boldsymbol{x}} = \boldsymbol{x} + \mathbf{n}$
    $\boldsymbol{x} = \lambda_p \Phi_p(\tilde{\boldsymbol{x}}) + (1 - \lambda_p)\tilde{\boldsymbol{x}}$
**end for**
$\boldsymbol{x}_{\text{aug}} = \text{RandomInvert}(\text{Normalize}(\text{Sigmoid}(\boldsymbol{x})))$
**return** $\boldsymbol{x}_{\text{aug}}$

## 4 Ensemble Pseudo-labeling

We introduce an ensembling method with random augmentation techniques to reinforce the reliability and robustness of pseudo-labels of DaPP. Specifically, given a dataset $\mathcal{D}_t$, we generate $R$ randomly augmented datasets $\{A_r\}_{r=1}^{R}$ and all $N_t \times r$ samples are aggregated to calculate prototypes. Then, pseudo-labels are calculated with an ensemble of predictions from different augmentations by $\hat{y}(\boldsymbol{x}_t) = \arg\max_k \sum_r \delta_k(-\operatorname{dist}(\bar{g}(A_r(\boldsymbol{x}_t)), C_{t,k}))$.



\end{document}

%% file: math_commands.tex
\usepackage{amsmath,amsfonts,bm}

\def\eqref#1{equation~\ref{#1}}

\def\1{\bm{1}}

\def\ru{{\textnormal{u}}}

\def\vx{{\bm{x}}}

\DeclareMathAlphabet{\mathsfit}{\encodingdefault}{\sfdefault}{m}{sl}
\SetMathAlphabet{\mathsfit}{bold}{\encodingdefault}{\sfdefault}{bx}{n}

\def\sR{{\mathbb{R}}}

\def\sX{{\mathbb{X}}}

\DeclareMathOperator*{\argmin}{arg\,min}